\documentclass[11pt]{article}
\usepackage[a4paper,margin=2.4cm]{geometry}
\usepackage[T1]{fontenc}
\usepackage{tgtermes,tgheros}
\usepackage{amsmath,amssymb}
\usepackage{graphicx}
\usepackage{placeins}
\usepackage{xcolor}
\usepackage[super,sort&compress,comma]{natbib}

\usepackage{setspace}
\usepackage[font=small,labelfont=bf]{caption}
\DeclareCaptionLabelSeparator{figurebar}{\space\textbar\space}
\renewcommand{\figurename}{Fig.}
\usepackage[hidelinks,pdfusetitle,pdfdisplaydoctitle]{hyperref}
\usepackage{xurl}


\title{Language models seek help despite guaranteed score losses in question answering}
\author{Jiahao You, Ziye Jia, Chao Dong, Fuhui Zhou, and Qihui Wu}
\date{}

\begin{document}
\maketitle

\begin{abstract}
\noindent At the highest waiting cost in a multiple-choice question-answering task, all 7 tested open-weight language models still consult an expert even though a correct expert answer earns fewer points than an incorrect independent answer. Pooling over expert accuracy and fixed-fee combinations gives help-seeking rates of 20.8\% to 94.1\% across models. We test 2,291 questions across 54 combinations of expert accuracy and consultation costs, comparing observed decisions with an expected-score baseline based on independently measured and calibrated model confidence. Help-seeking rates vary across conditions with the same expected consultation score. Fine-tuning reduces decision regret by 59\% to 87\% on held-out questions within the training range but increases inconsistency at matched expected consultation scores. In a separate task, external routing improves scores by 0.180 to 0.273 points per question with candidate answers held fixed.
\end{abstract}

\section*{Introduction}

Metacognition allows individuals to assess their own judgments and regulate subsequent decisions. Confidence helps humans and other animals determine whether to act on a judgment\cite{kepecs2008confidence,kiani2009confidence,fleming2017metacognition}. In humans, lower confidence predicts additional information seeking even when objective accuracy is matched\cite{desender2018information}. Advice-seeking experiments examine this relationship when participants can pay for advice\cite{pescetelli2021advice}. The value of additional information depends on its expected benefits and the time and effort required to obtain it\cite{petitet2021sampling,schulz2023metacognitive}. Consulting an expert involves choosing between an initial judgment and an answer from an external source. When the expert's answer is used directly, this choice depends on confidence in the initial answer, the expert's reliability and the consultation cost.

Large language model (LLM) agents face a similar decision when answering independently or consulting tools, other models or people\cite{schick2023toolformer,ren2023robots}. Agents also overuse tools on tasks they can solve with their own knowledge\cite{qian2025smart}. Studies examine how well model confidence reflects answer correctness and how models communicate uncertainty to users\cite{xiong2024uncertainty,tian2023calibration,steyvers2025know}. Causal evidence shows that confidence-related signals influence abstention, and that models adjust their behavior to instructed confidence thresholds\cite{kumaran2026causal}. Work on LLM escalation also reports model-dependent responses to decision costs and improvements from fine-tuning\cite{disorbo2026escalation}. We examine whether confidence-related help-seeking reliably tracks the relative expected scores of independent answering and consultation.

Selective prediction and learning-to-defer methods compare the expected outcomes of available actions\cite{chow1970optimum,madras2018predict,mozannar2020consistent,narasimhan2022posthoc}. In our consultation task, confidence estimates the chance of success without help, while stated expert accuracy supplies an estimate for consultation. Under an expected-score objective, consultation is worthwhile when the expected improvement from switching answer sources exceeds the consultation cost. For a fixed question and confidence estimate, we test whether choices follow changes in expected scores, remain consistent across conditions with the same expected consultation score, and avoid consultation when it guarantees a lower score.

We study 7 open-weight models from 3 developers on the same questions across 54 combinations of expert accuracy, fixed fees and waiting costs. We measure and calibrate confidence using a separate answer-only prompt and keep these estimates fixed while varying the consultation conditions. This provides an expected-score baseline for evaluating responses to expert accuracy and costs. Some conditions make even a correct expert answer score below an incorrect independent answer, giving a test determined entirely by the scoring rules. We compare help-seeking when points are deducted for consulting an expert and when points are deducted for answering independently. We also test how changes in prompt wording affect help-seeking. Numerical tasks measure how accurately models compare stated values and select the action with the higher expected score. Using activation patching, we replace internal states with those from another condition for the same question and measure changes in consultation decisions. We introduce confidence-anchored payoff training (CAPT) and compare 3 training objectives using decision regret and consistency at matched expected consultation scores. For external routing, we hold model and expert answers fixed and use an expected-score rule to choose between them.

Help-seeking persists in all 7 models when consultation guarantees a lower score. At the highest waiting cost, pooled help-seeking rates range from 20.8\% to 94.1\%. Help-seeking rates also vary across conditions with the same expected consultation score. In numerical tasks, all 7 models compare stated numbers accurately, yet they select the higher-scoring action on only 58.5\% to 69.0\% of trials when given accuracy and cost values. Providing both expected scores raises this to 83.4\% to 99.7\% but reduces independent-answer accuracy in question answering. Across the 3 tested training objectives, fine-tuning reduces decision regret by 59\% to 87\% on held-out questions within the training range and increases decision inconsistency at matched expected consultation scores. In a task with an expert model, external routing that selects between fixed candidate answers improves scores by 0.180 to 0.273 points per question. These findings can guide the evaluation and design of language model agents that must balance the benefits of external assistance against its costs.

\section*{Results}

\subsection*{Help-seeking persists under guaranteed score losses}

\begin{figure}[!tb]
\captionsetup{labelfont={bf,color=black}}
\centering
\makebox[\textwidth][c]{\includegraphics[width=160mm]{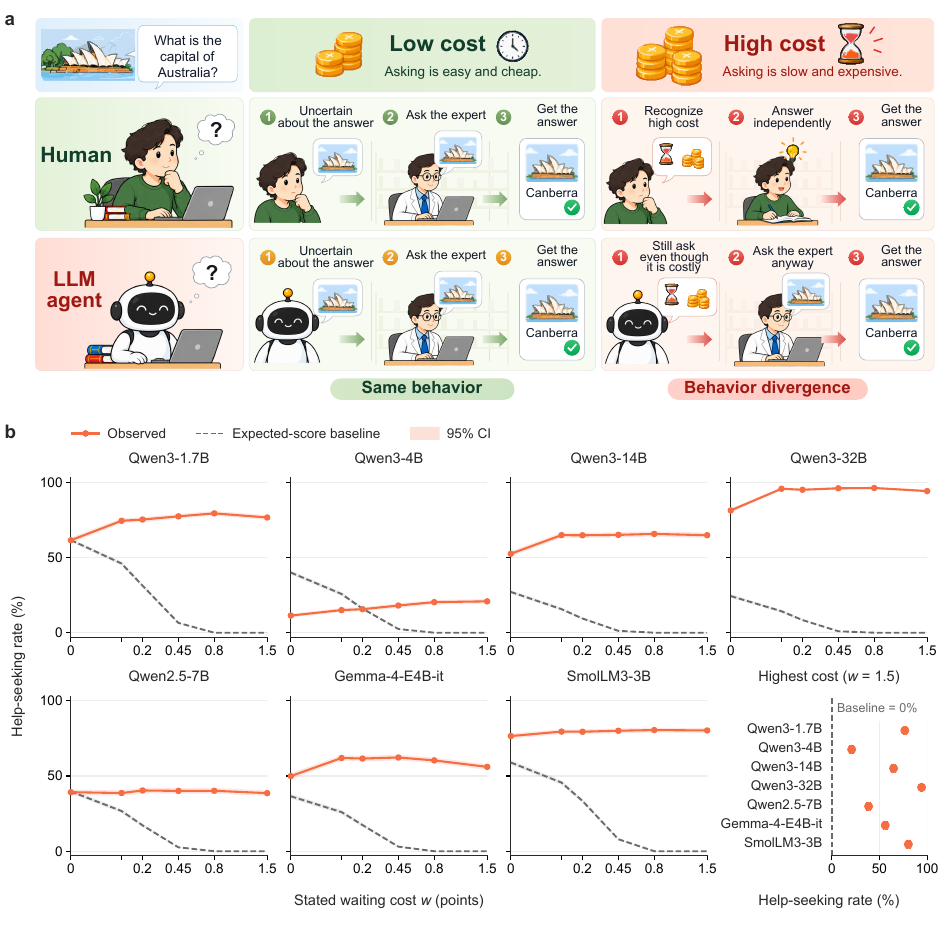}}
\caption{\textbf{Help-seeking persists when consultation guarantees a score loss.} \textbf{a} Conceptual responses to consultation costs. \textbf{b} Help-seeking rates for the 7 models pooled over 9 expert-accuracy and fixed-fee combinations. Orange denotes observations and gray dashed lines the confidence-based expected-score baseline. The waiting-cost axis uses a square-root scale, with $w=1.5$ summarized at lower right. Shading and bars show the 95\% paired bootstrap confidence interval (CI) over 2,291 questions for each model.}
\label{fig:task}
\end{figure}

All 7 models in Fig.~\ref{fig:task} continue to seek help when consultation guarantees a lower score than independent answering. Each model chooses between independent answering and consultation on 2,291 4-option MMLU-Redux questions\cite{gema2025mmlu} across 54 accuracy-cost combinations. Correct answers earn 1 point regardless of their source, while consultation incurs a fixed fee and a waiting cost. At the highest waiting cost of 1.5 points, even a correct expert answer earns at most $-0.5$ points, while independent answering earns 0 or 1 point. Independent answering therefore scores higher for every possible outcome. Pooled over 9 accuracy-fee combinations, observed rates at the highest cost range from 20.8\% to 94.1\%. The corresponding 95\% bootstrap confidence intervals are [19.3\%, 22.3\%] and [93.3\%, 95.0\%], respectively.

\begin{figure}[!tb]
\captionsetup{labelfont={bf,color=black}}
\centering
\makebox[\textwidth][c]{\includegraphics[width=180mm]{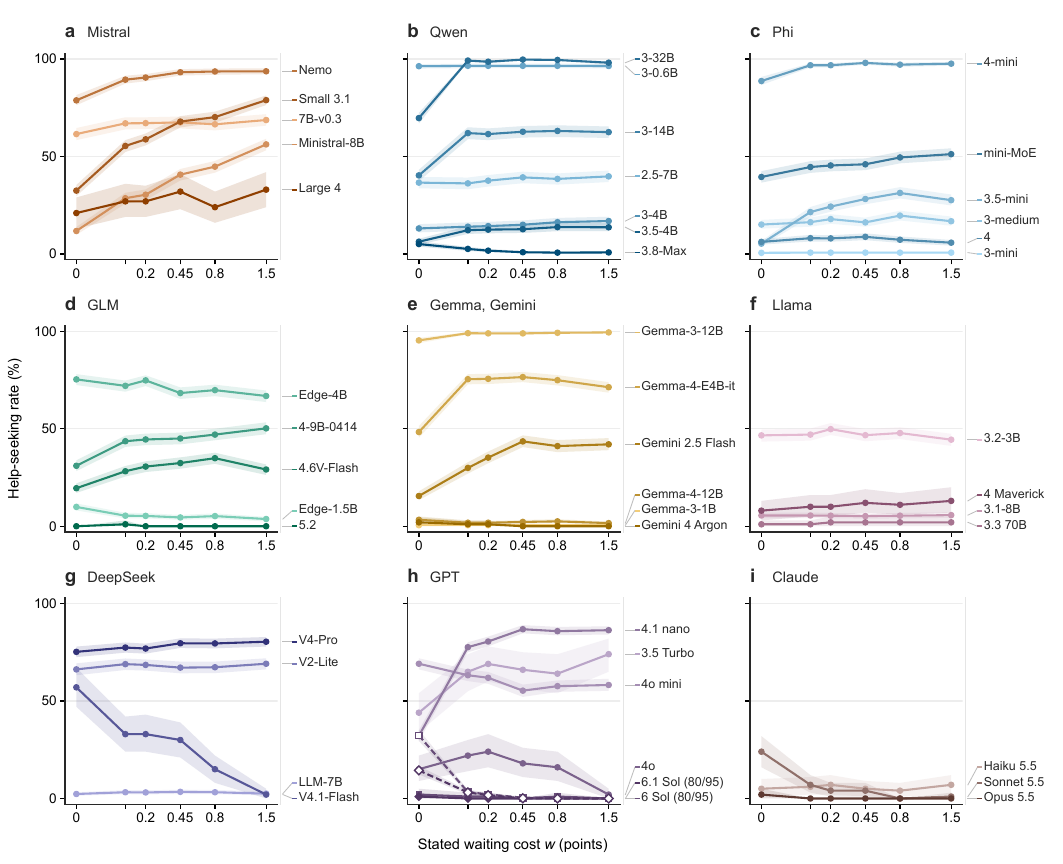}}
\caption{\textbf{Variation in help-seeking across models.} Responses to 6 waiting costs on MMLU-Redux with no fixed fee. Panels \textbf{a--i} show 46 models from 9 developers. Solid lines and filled markers indicate $a_H=0.80$. Dashed lines and open markers indicate $a_H=0.95$ for GPT-6 Sol and GPT-6.1 Sol. Colors distinguish families and shades distinguish models. The waiting-cost axis uses a square-root scale. Bands show pointwise 95\% bootstrap confidence intervals. Each curve uses 100 or 1,000 questions per cost, with response formats varying by model. For GPT-6 Sol and GPT-6.1 Sol, the 80\% and 95\% curves use unmatched sets of 100 and 1,000 questions, respectively.}
\label{fig:coverage}
\end{figure}

Introducing a small consultation cost increases help-seeking in several models. The baseline implies a reduction of 10.1 to 15.6 percentage points when the waiting cost rises from 0 to 0.1 points. Observed changes instead range from $-0.6$ to +14.4 percentage points. Introducing this first cost raises the help-seeking rate by 12.0 to 14.4 percentage points in Qwen3-1.7B, Qwen3-14B, Qwen3-32B and Gemma-4-E4B-it. A total of 6 models increase asking or show little change across both cost contrasts. Gemma-4-E4B-it reduces asking from 62.0\% to 56.0\% as the waiting cost rises from 0.1 to 1.5 points, but still asks on more than 50\% of trials when every request loses points. At a waiting cost of 0.8 points, where the baseline help-seeking rate is 0\% in every model, observed rates range from 20.3\% to 96.2\%. Relative to 0 waiting cost, the help-seeking rate at the highest cost rises by 3.7 to 15.1 percentage points in 6 models. It falls by 0.6 percentage points in Qwen2.5-7B-Instruct. The experiment with 46 models on MMLU-Redux in Fig.~\ref{fig:coverage} shows varied responses to consultation costs. At the highest waiting cost, the lower 95\% confidence bound of the help-seeking rate exceeds 5\% in 28 of the 46 displayed models and 50\% in 16 models.

\begin{figure}[!tb]
\captionsetup{labelfont={bf,color=black}}
\centering
\makebox[\textwidth][c]{\includegraphics[width=160mm]{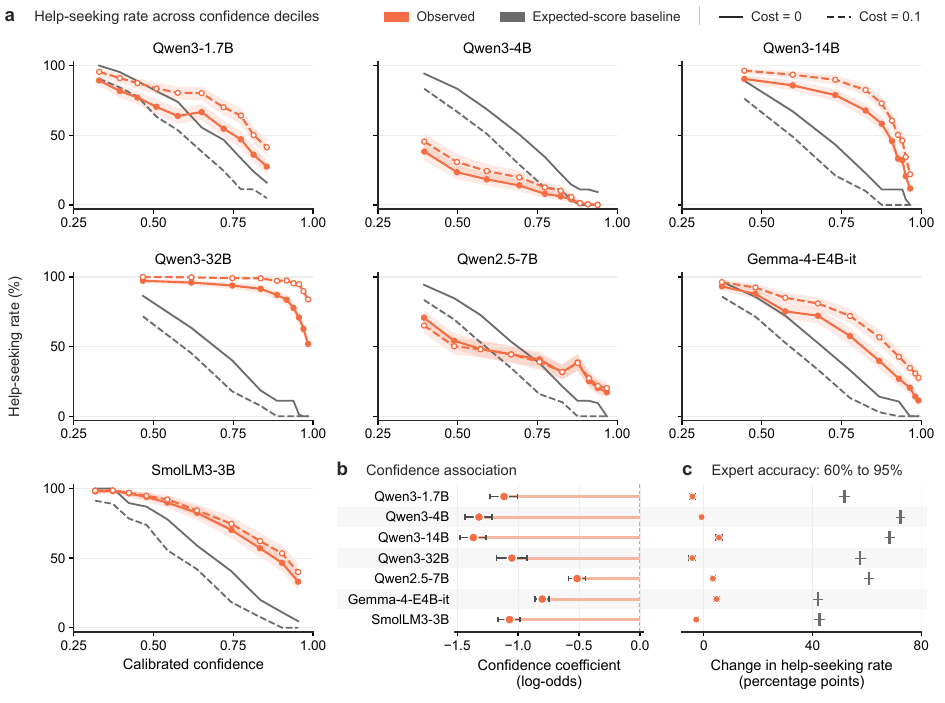}}
\caption{\textbf{Help-seeking is associated with lower confidence and responds weakly to expert accuracy.} \textbf{a} Help-seeking across confidence deciles in 7 models, at waiting costs of 0 and 0.1 points, shown by solid and dashed lines, respectively. \textbf{b} Logistic-regression coefficients for confidence, adjusted for 54 conditions. \textbf{c} Responses to expert accuracy increasing from 60\% to 95\% without costs. Orange denotes observations and gray the expected-score baseline. Shading and error bars show 95\% bootstrap confidence intervals over 2,291 questions per model, with fixed bins in \textbf{a} and paired questions in \textbf{c}.}
\label{fig:confidence}
\end{figure}

The results in Figs.~\ref{fig:confidence}a and b show that lower calibrated confidence is associated with more frequent help-seeking in all 7 models. At 0 waiting cost, the help-seeking rate ranges from 38.3\% to 97.9\% in the least confident decile and from 0.0\% to 52.0\% in the most confident decile. Nevertheless, raising the waiting cost increases observed asking in every confidence group in 5 models, whereas the expected-score baseline implies lower or unchanged help-seeking rates. In the most confident decile at 0.1 points, Qwen3-32B asks on 83.9\% of trials, whereas the baseline implies at most 4.9\% in any model. After adjustment for the 54 conditions, the logistic-regression coefficients range from $-1.367$ to $-0.516$, with all 7 95\% bootstrap confidence intervals below 0.

In Fig.~\ref{fig:confidence}c, increasing the expert's stated accuracy from 60\% to 95\% produces small or negative changes in help-seeking when both the fixed fee and waiting cost are 0. The baseline implies an increase in asking of 42.0 to 72.4 percentage points. Observed changes instead range from $-4.3$ percentage points, with a 95\% bootstrap confidence interval of [$-5.6$, $-3.1$], to +5.6 percentage points, with an interval of [+4.4, +6.8]. Qwen3-1.7B, Qwen3-4B, Qwen3-32B and SmolLM3-3B reduce asking by 0.8 to 4.3 percentage points despite the higher stated expert accuracy. Among models with increased help-seeking, the observed increase is at most about 11\% of the increase implied by the baseline.

\begin{figure}[!tb]
\captionsetup{labelfont={bf,color=black}}
\centering
\makebox[\textwidth][c]{\includegraphics[width=180mm]{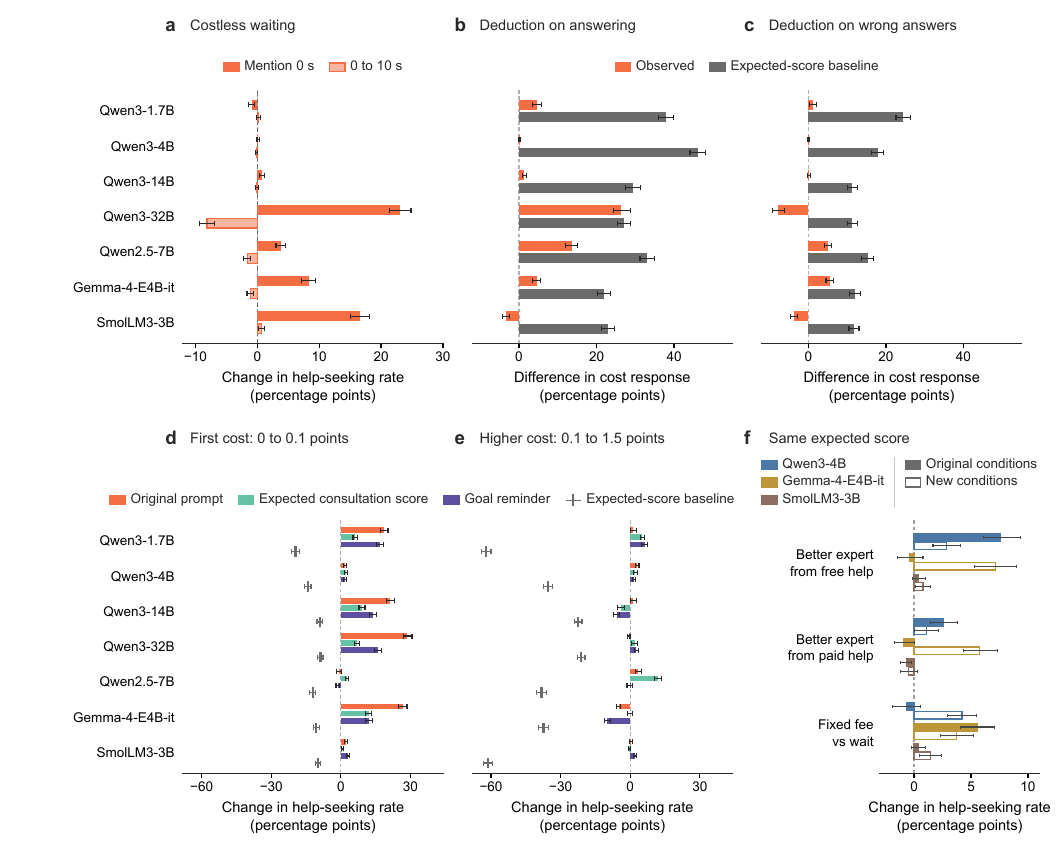}}
\caption{\textbf{Cost wording and scoring consequences affect help-seeking.} \textbf{a} Responses to mentioning or extending a wait without point deductions. \textbf{b,c} Cost responses when deductions move to independent answering or incorrect independent answers. \textbf{d,e} Responses to costs increasing from 0 to 0.1 and from 0.1 to 1.5 points under 3 prompts. \textbf{f} Conditions with the same expected consultation score in 3 models. Filled and open bars denote the original setting and a new setting with changed expert accuracy, costs and wording. Gray lines show the expected-score baseline. Error bars show 95\% bootstrap confidence intervals over questions. Each model uses 2,291 questions at 80\% expert accuracy in \textbf{a--e}, and 512 ARC-Challenge questions with 2 label orders in \textbf{f}.}
\label{fig:cost}
\end{figure}

\subsection*{Help-seeking responses to costs and score information}

Help-seeking in Fig.~\ref{fig:cost} varies with waiting statements, deduction rules and prompt wording. In Fig.~\ref{fig:cost}a, with no points deducted, adding that the expert replies after 0 seconds increases asking by 0.7 to 23.0 percentage points in 5 of 7 models. Extending the wait to 10 seconds reduces asking by 8.1, 1.6 and 1.1 percentage points in Qwen3-32B, Qwen2.5-7B-Instruct and Gemma-4-E4B-it, respectively. In SmolLM3-3B, asking increases by 0.7 percentage points. The response to introducing a cost in Fig.~\ref{fig:cost}b changes less than the baseline implies in 6 models when deductions move from consultation to independent answering. Qwen3-32B is close to the baseline. Applying the deduction only to incorrect independent answers also changes the response by less than the baseline implies. In Fig.~\ref{fig:cost}c, Qwen3-32B and SmolLM3-3B respond in the direction opposite to that implied by the baseline.

When the expected consultation score is supplied directly, introducing the first cost increases help-seeking by 0.7 to 12.0 percentage points across the 7 models. The 95\% intervals at the 2 extremes are [0.3, 1.1] and [10.8, 13.4]. With further increases, asking falls in Qwen3-14B, while the other models show small or positive changes. With a reminder to maximize expected score, Gemma-4-E4B-it increases asking by 12.0 percentage points at the first cost step. It reduces asking by 9.8 percentage points at the higher step, as shown in Figs.~\ref{fig:cost}d and e.

\begin{figure}[!tb]
\captionsetup{labelfont={bf,color=black}}
\centering
\makebox[\textwidth][c]{\includegraphics[width=162mm]{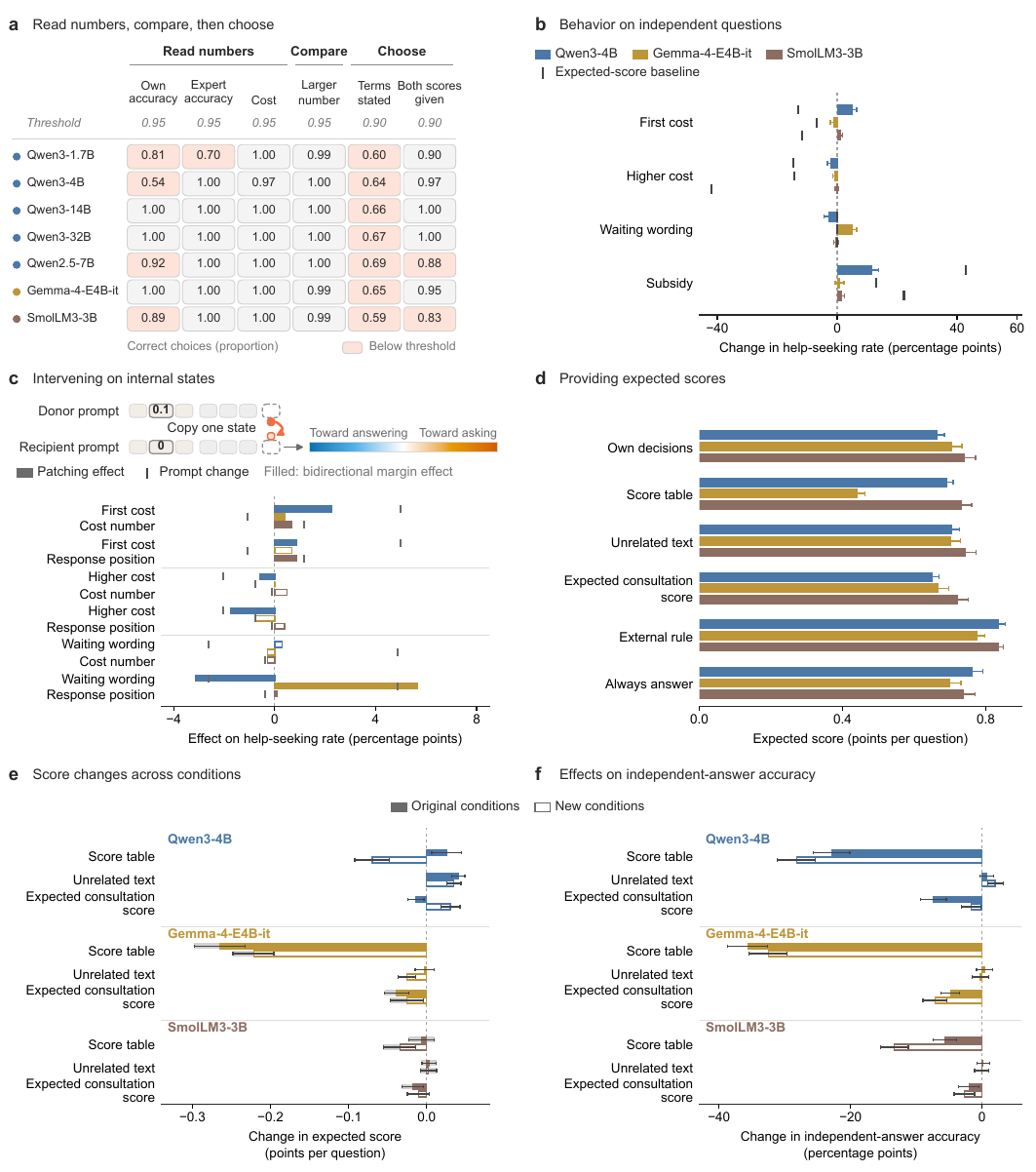}}
\caption{\textbf{Providing expected scores improves numerical action selection and reduces independent-answer accuracy in question answering.} \textbf{a} Accuracy on numerical tasks. Shading marks point estimates below task thresholds. \textbf{b} Help-seeking contrasts and the expected-score baseline. \textbf{c} Control-adjusted activation-patching effects on help-seeking rate, alongside prompt effects. Filled bars mark margin effects in both predicted directions, using two-sided sign-flip tests with Holm-adjusted $P\leq0.05$. \textbf{d} Expected scores under the original setting. Score tables provide confidence and expected consultation score. \textbf{e,f} Changes in score and independent-answer accuracy from unmodified prompts. Filled and open bars denote original and new settings. Panel \textbf{a} uses 400 numerical combinations per model and task. Panels \textbf{b--f} use 512 questions per model, averaging 36 conditions in \textbf{d--f}. Error bars show 95\% bootstrap confidence intervals. Gray intervals in \textbf{e} bound score changes when decision measurements are invalid.}
\label{fig:inside}
\end{figure}

On ARC-Challenge questions, help-seeking rates in Fig.~\ref{fig:cost}f vary across conditions with the same expected consultation score. Increasing expert accuracy and cost by the same amount, starting from either free or paid help, changes help-seeking in some models and conditions. Replacing a waiting cost with an equal fixed fee also changes help-seeking. For example, raising accuracy and cost from free help increases asking in Qwen3-4B by 7.6 percentage points under the original setting, with a 95\% interval of [6.1, 9.4].

In the numerical task in Fig.~\ref{fig:inside}a, providing both expected scores improves action selection across all 7 models. A total of 6 models identify the expert's stated accuracy correctly on every trial, and all 7 identify the cost with 97.4\% to 100\% accuracy. Accuracy in identifying the probability of answering correctly given in the prompt ranges from 53.7\% to 100\%. All 7 models identify the larger of 2 numbers with 98.5\% to 100\% accuracy. When given accuracy and cost values, they select the action with the higher expected score on 58.5\% to 69.0\% of trials. Supplying both expected scores raises this accuracy to 83.4\% to 99.7\%, with the lower confidence bound exceeding 90\% in Qwen3-4B, Qwen3-14B, Qwen3-32B and Gemma-4-E4B-it.

We test Qwen3-4B, SmolLM3-3B and Gemma-4-E4B-it on ARC-Challenge questions under 2 settings. Under the original setting in Fig.~\ref{fig:inside}b, 2 models increase help-seeking at low costs, and all 3 continue to consult under guaranteed losses. Under the expected-score baseline, raising the stated cost from 0 to 0.1 points implies reductions in asking of 12.9, 11.7 and 6.8 percentage points in Qwen3-4B, SmolLM3-3B and Gemma-4-E4B-it. Instead, asking rises by 5.0 percentage points in Qwen3-4B, with an interval of [3.6, 6.4], and by 1.2 percentage points in SmolLM3-3B, with an interval of [0.6, 1.9]. Gemma-4-E4B-it changes by $-1.1$ percentage points, with a 95\% confidence interval of [$-2.2$, +0.1]. Under the prompt supplying the expected consultation score, the same cost step increases help-seeking in Fig.~\ref{fig:cost}d. Raising the cost to 1.5 points, where every request loses points, lowers asking by at most 2.1 percentage points, against decreases of 14.5 to 41.8 percentage points implied by the baseline. At that cost and 80\% stated expert accuracy, Qwen3-4B, Gemma-4-E4B-it and SmolLM3-3B still ask on 58.3\%, 12.3\% and 6.9\% of the ARC-Challenge questions, respectively. Wording changes that leave the expected-score baseline unchanged also alter help-seeking. Stating the cost as a wait changes asking by $-2.6$ percentage points in Qwen3-4B, with an interval of [$-4.3$, $-1.1$], and by +4.9 percentage points in Gemma-4-E4B-it, with an interval of [+3.2, +6.4].

Activation patching in Fig.~\ref{fig:inside}c shifts decision margins in the directions associated with selected changes in costs and wording. The effect of introducing a cost on the decision margin is reproduced at both positions in Qwen3-4B and SmolLM3-3B. The effect of waiting-cost wording is reproduced at the response position in all 3 models, but not at the cost-number position. Across the original and new settings, we evaluate 36 activation-patching comparisons, as shown in Supplementary Fig.~\ref{supp:swaps}. In 18 comparisons, both forward and reverse interventions produce significant changes in the decision margin in the predicted directions after correction for multiple comparisons.

In Figs.~\ref{fig:inside}d--f, adding a score table lowers independent-answer accuracy in the ARC-Challenge task. Under the original setting, the score table changes expected score by $-0.265$ to +0.026 points per question and lowers independent-answer accuracy by 5.6 to 35.6 percentage points. Qwen3-4B's small score gain is not statistically significant after correction for multiple comparisons. Under the new setting, the table lowers both score and independent-answer accuracy in all 3 models. Giving only the expected consultation score also reduces independent-answer accuracy in every model and setting, whereas unrelated text produces much smaller accuracy changes. In Fig.~\ref{fig:inside}d, selecting the answer source with the external expected-score rule improves scores for all 3 models relative to their own decisions under the original prompt.

\subsection*{Training and external routing of help-seeking}

\begin{figure}[!tb]
\captionsetup{labelfont={bf,color=black}}
\centering
\makebox[\textwidth][c]{\includegraphics[width=180mm]{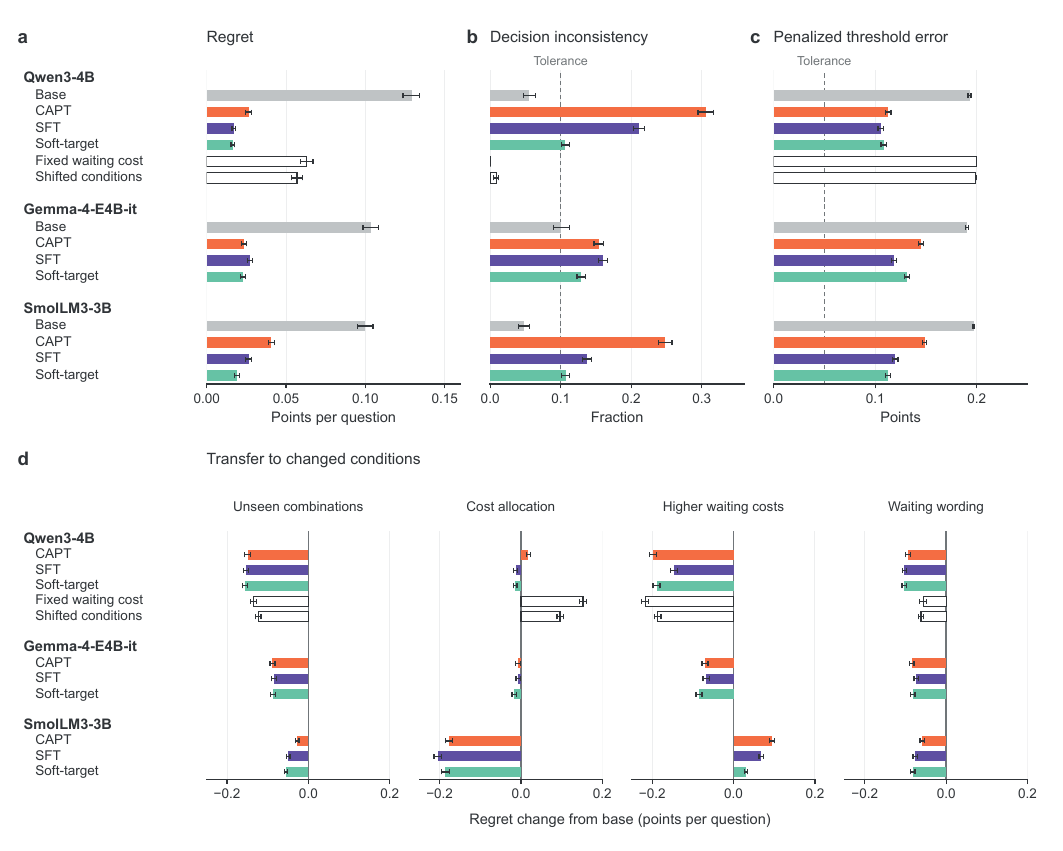}}
\caption{\textbf{Fine-tuning reduces regret and increases inconsistency at matched expected consultation scores.} \textbf{a} Regret on held-out questions within the training range. \textbf{b} Decision inconsistency across 3 conditions with the same expected consultation score, measured by $E_3$. \textbf{c} Penalized threshold error $E_1$, including a 0.20 endpoint penalty. \textbf{d} Regret changes under 4 new settings, with negative values indicating improvement. Open bars denote controls and dashed lines evaluation tolerances. Each model uses 2,000 questions, 3 seeds per main objective and 1 per control. Intervals are 95\% bootstrap confidence intervals after averaging across seeds within each question.}
\label{fig:training}
\end{figure}

Fine-tuning reduces regret across the models and objectives tested in Fig.~\ref{fig:training}. We compare CAPT, supervised fine-tuning (SFT) and soft-target training. Across 9 combinations of models and training objectives, training reduces regret by 59\% to 87\% on held-out questions within the training range. Regret falls from 0.100--0.129 points per question in the base models to 0.016--0.041 after training. Qwen3-4B's controls with a fixed waiting cost or cyclically shifted conditions also reduce regret by 51\% and 56\%, compared with 80\% for CAPT in Fig.~\ref{fig:training}a. The main objectives change answer-only accuracy on the evaluation questions by $-0.9$ to +1.0 percentage points and expected calibration error by $-0.020$ to +0.034.

The main training objectives increase decision inconsistency from 0.05--0.10 in the base models to 0.11--0.31. In Figs.~\ref{fig:training}b and c, penalized threshold error falls from 0.19--0.20 to 0.11--0.15. Decisions stay close to a single threshold rule before and after training. Disagreement with the fitted rule, $E_2$, is 0.004--0.019 in the base models and at most 0.003 after training. Here, $E_2$ measures threshold fit within the 8-point scan, while $E_3$ compares decisions across a separate set of conditions with matched expected consultation scores. Among the main objectives, soft-target training gives the lowest inconsistency in each model, from 0.107 to 0.129. The controls with a fixed waiting cost or cyclically shifted conditions have inconsistency below 0.01. Their penalized threshold errors of 0.200 and 0.199 approach the 0.20 penalty assigned when no interior threshold is fitted.

Across all 9 model-objective combinations, regret falls by 0.027 to 0.155 points per question for combinations of expert accuracy and costs not used in training. Under waiting-cost wording in Fig.~\ref{fig:training}d, it falls by 0.059 to 0.103. When deductions apply to independent answering or only to incorrect independent answers, training reduces regret by 0.177 to 0.205 points per question in SmolLM3-3B. Training raises regret by 0.152 and 0.096 in Qwen3-4B's controls with a fixed waiting cost or cyclically shifted conditions, respectively. At higher waiting costs of 0.4 to 0.7 points, the main objectives reduce regret by 0.067--0.198 points per question in Qwen3-4B and Gemma-4-E4B-it. The Qwen3-4B controls with a fixed waiting cost or cyclically shifted conditions reduce it by 0.217 and 0.187. Regret in SmolLM3-3B rises by 0.031--0.095 in Fig.~\ref{fig:training}d.

\begin{figure}[!tb]
\captionsetup{labelfont={bf,color=black}}
\centering
\makebox[\textwidth][c]{\includegraphics[width=180mm]{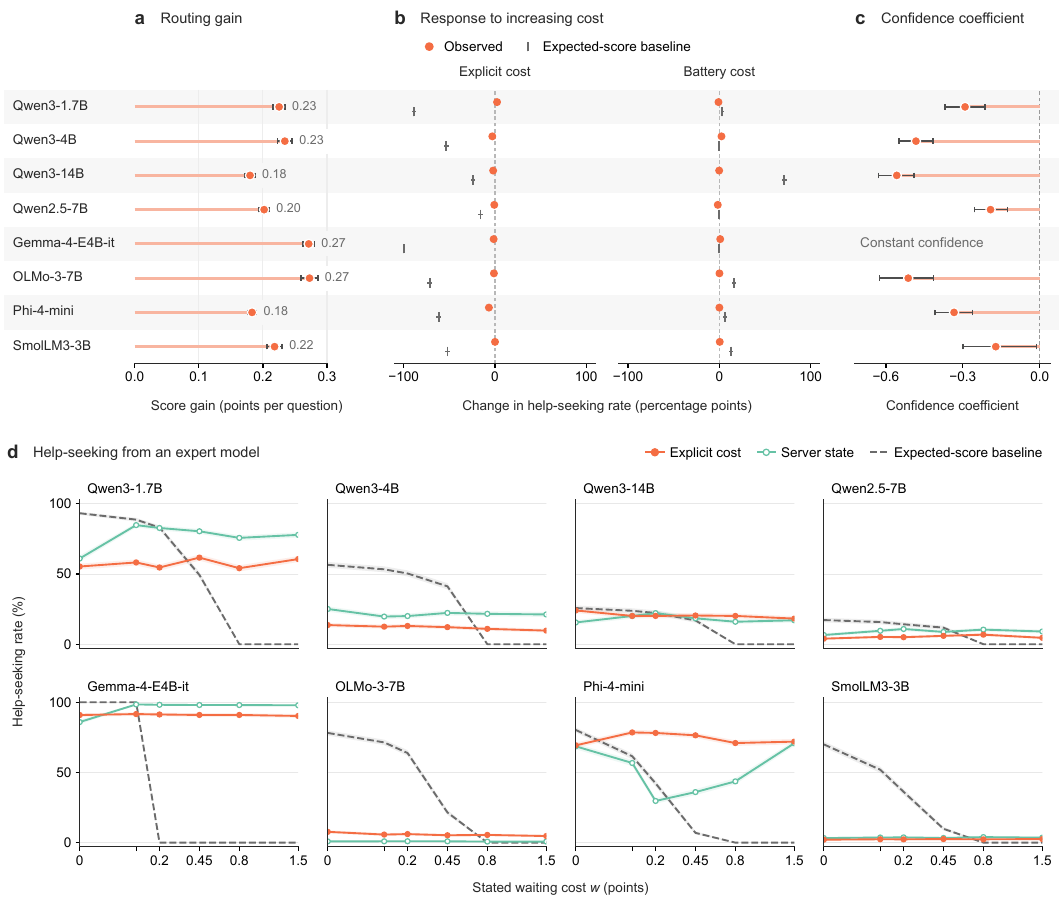}}
\caption{\textbf{External expected-score routing improves scores with fixed candidate answers.} \textbf{a} Routing gains in 8 models with fixed assistant and expert answers, averaged over 18 conditions. \textbf{b} Responses to costs increasing from 0.1 to 1.5 points, charged to consultation or to battery use during independent answering. \textbf{c} Logistic-regression coefficients for confidence, adjusted for condition. Gemma-4-E4B-it has constant confidence of 0.516 and no slope estimate. \textbf{d} Help-seeking under explicit costs and server-state descriptions. Gray lines show the expected-score baseline. The cost axis uses a square-root scale. Each model uses 2,000 questions. Error bars and bands show 95\% bootstrap confidence intervals over questions.}
\label{fig:routing}
\end{figure}

The external expected-score rule in Fig.~\ref{fig:routing}a improves routing in the task with an expert model. A total of 8 models from 5 developers answer 2,000 held-out TriviaQA questions\cite{joshi2017triviaqa} or request an answer from Qwen3-32B. We measure calibrated verbalized confidence. With model and expert answers held fixed, the external rule earns 0.180--0.273 more points per question than each model's own routing, averaged over 18 conditions. The gain is largest in OLMo-3-7B-Instruct and Gemma-4-E4B-it, at 0.273 and 0.271 points per question, and smallest in Qwen3-14B, at 0.180. A routing rule using development-set answer accuracy instead of question-level confidence yields gains of 0.185--0.276 points per question. The mean score difference between question-level confidence routing and this control is $-0.011$ to 0.000 points per question. Routing gains remain positive under all 3 alternative answer-matching rules, with all 95\% bootstrap intervals above 0.

In Fig.~\ref{fig:routing}b, responses to explicit consultation costs remain much smaller than those implied by the expected-score baseline. As the cost rises from 0.1 to 1.5 points, observed changes range from $-6.6$ to +2.4 percentage points, whereas the baseline implies decreases of 15.8 to 100 percentage points. Under battery costs charged to independent answering, help-seeking increases by 2.4 and 1.0 percentage points in Qwen3-4B and Gemma-4-E4B-it. It decreases by 1.7 percentage points in Qwen2.5-7B-Instruct. The baseline help-seeking rate remains unchanged in all 3 models. In Qwen3-14B, the baseline implies an increase of 71.4 percentage points under battery costs, whereas the observed change is $-0.2$ percentage points. Among the 7 models with varying calibrated confidence in Fig.~\ref{fig:routing}c, lower confidence is associated with more frequent help-seeking. Their coefficients range from $-0.558$ to $-0.171$, and all 7 95\% bootstrap confidence intervals are below 0. In Fig.~\ref{fig:routing}d, 5 models still request help on more than 5\% of questions when consultation loses at least 0.5 points, including 90.3\% in Gemma-4-E4B-it and 72.0\% in Phi-4-mini-instruct. When the same delays are described as server states, asking in Phi-4-mini-instruct falls from 68.6\% to 29.7\% as the delay grows to 20 seconds and returns to 70.8\% at 150 seconds.

\section*{Discussion}

Our findings identify failures in how language models select external assistance under explicit scoring rules. All 7 models in the main experiment seek help when even a correct expert answer scores below an incorrect independent answer. The resulting loss is established for every possible answer outcome and does not depend on estimating the model's confidence or the expert's accuracy. Help-seeking also changes when accuracy and costs vary while the expected consultation score remains fixed. Together, these tests assess whether models avoid guaranteed losses and choose consistently across conditions with the same expected consultation score. They provide concrete criteria for evaluating cost-sensitive help-seeking under an expected-score objective.

Our study extends works on confidence-guided abstention to decisions involving external answers with varying reliability and cost. Kumaran and colleagues show that confidence-related signals influence abstention and that models adjust to instructed confidence thresholds\cite{kumaran2026causal}. In our experiments, lower calibrated confidence is associated with more frequent help-seeking after adjustment for 54 accuracy-cost conditions, whereas increasing stated expert accuracy produces small or negative changes. Therefore, confidence-related help-seeking coexists with weak responses to the reliability of the alternative answer source. Selecting assistance requires considering both answer sources and the cost of switching between them, placing demands on consultation decisions beyond their association with confidence. Previous works identify asymmetric responses to supporting and opposing advice after it is received\cite{competing2026biases}. Our results locate additional decision errors in the earlier choice of whether to obtain that advice.

The behavioral controls show how consultation decisions respond to both scoring consequences and their presentation. Introducing a consultation cost increases help-seeking in several models. Stating a waiting time without deducting points also increases help-seeking in several models. Changes in wording and exchanges between equal waiting costs and fixed fees also alter decisions. This sensitivity is consistent with reports of framing and wording effects in language model decisions\cite{chen2023rationality,binz2023cognitive,echterhoff2024bias}. In separate numerical tasks, all 7 models compare stated numbers accurately, and providing both expected scores improves action selection. However, in question answering, providing expected scores reduces independent-answer accuracy, whereas unrelated text produces much smaller changes. In these experiments, score information affects both consultation decisions and the answers generated for independent answering. Activation patching provides intervention evidence linking selected internal states to these behavioral responses: replacing states between conditions for the same question shifts decision margins in directions associated with cost and wording changes. Together, these experiments show why evaluating help-seeking requires examining the relationships among numerical processing, answer generation and answer-source selection.

Training reveals that lower decision regret can coexist with greater inconsistency at matched expected consultation scores. We introduce CAPT to train consultation decisions against expected action scores anchored to frozen model confidence, while constraining changes in answer distributions. Across CAPT, supervised fine-tuning and soft-target training, regret falls by 59\% to 87\% on held-out questions within the training range. Penalized threshold error also decreases, and decisions fit a threshold rule more closely. Yet decision inconsistency increases across a separate set of conditions with the same expected consultation score. Qwen3-4B's controls with a fixed waiting cost or cyclically shifted conditions also reduce regret, while penalized threshold errors remain at 0.200 and 0.199, near the endpoint penalty. Prior work improves escalation and tool-calling decisions through training on uncertainty and costs\cite{disorbo2026escalation,xu2025toolcalling}. Our results show why such improvements should be evaluated jointly with consistency across matched-score conditions and threshold behavior. Future studies could add a training penalty for inconsistent choices across conditions with the same expected consultation score. They could then test whether this improves consistency while preserving the reduction in decision regret.

External routing demonstrates a practical gain from improving answer-source selection while keeping answer content fixed. Across 8 assistant models in the expert-consultation task, an expected-score rule improves scores by 0.180 to 0.273 points per question, averaged over 18 conditions. Because the assistant and expert candidate answers are identical across routing policies, these gains are attributable to which source is selected. A rule using development-set accuracy achieves gains comparable to those from question-level confidence, showing that a single accuracy estimate per model supports effective routing in these experiments. Building on selective prediction, learning to defer, model cascades and routing\cite{chow1970optimum,kamath2020selective,wen2025know,mozannar2020consistent,narasimhan2022posthoc,jitkrittum2023cascade,ong2025routellm}, this comparison quantifies the score improvement available from changing the selection rule alone. It also avoids the changes to independent answers observed when score information is added to the answering prompt. With candidate answers fixed, applying the expected-score comparison outside answer generation recovers gains left unrealized by the models' own consultation decisions.

\section*{Methods}

\subsection*{Models, datasets and confidence measurement}

We evaluate 7 open-weight language models: Qwen3-1.7B, Qwen3-4B, Qwen3-14B, Qwen3-32B, Qwen2.5-7B-Instruct, Gemma-4-E4B-it and SmolLM3-3B. Qwen3-14B uses online 8-bit floating-point (FP8) quantization. Qwen3-32B uses 4-bit weights produced by activation-aware weight quantization (AWQ)~\cite{lin2025awq}. The remaining models use bfloat16 (bf16) precision. The main experiments use vLLM~\cite{kwon2023vllm} with greedy decoding, whereas activation-patching experiments use Hugging Face Transformers in bf16. Reasoning mode is disabled in both settings. We select models for the main experiments only if their answer and consultation measurements are valid. We also require their help-seeking rate to be between 5\% and 95\% when stated expert accuracy is 80\% and consultation is free. MMLU-Redux~\cite{gema2025mmlu} covers multiple-choice questions across science, the humanities and social sciences. We sample 1,000 calibration questions and 2,000 evaluation questions by subject. The main help-seeking experiments use the remaining 2,291 questions annotated as error-free. TriviaQA~\cite{joshi2017triviaqa} contains open-domain questions with predominantly short answers. We use 2,000 questions without supporting context to study expert-model consultation. ARC-Challenge~\cite{clark2018arc} contains multiple-choice science questions. We use 512 4-option questions for behavioral experiments and activation patching.

We measure help-seeking decisions from token probabilities at the first response position. For question $i$ under condition $k$, $P_{ik}(j)$ denotes the probability of response digit $j$, with digits 1--4 representing the answer options and digit 5 representing consultation. The decision margin $m_{ik}$ is defined as
\begin{equation}
m_{ik}=\log P_{ik}(5)-\max_{1\leq j\leq4}\log P_{ik}(j),
\label{eq:readout}
\end{equation}
and the consultation indicator $Y_{ik}$ as
\begin{equation}
Y_{ik}=\mathbf{1}\{m_{ik}>0\}.
\label{eq:asking}
\end{equation}
When $Y_{ik}=0$, the model selects the most probable answer option. A decision measurement is valid when the allowed response digits carry at least 0.5 of the next-token probability. A condition meets the measurement-validity criterion when at least 99\% of its trials have valid decisions. Main-experiment and cost-control help-seeking rates include all trials.

Answer confidence is measured using an answer-only prompt without expert information or consultation terms, as illustrated in Supplementary Fig.~\ref{supp:prompts}. Following Kumaran and colleagues~\cite{kumaran2026causal}, we extract answer-option log-probabilities $s_{ij}$ and apply temperature scaling~\cite{guo2017calibration}. A model-specific temperature $T_c$ is fitted by minimizing the negative log-likelihood of correct answers on calibration questions. Calibrated confidence is defined as
\begin{equation}
z_i=\max_{1\leq j\leq4}
\frac{\exp(s_{ij}/T_c)}
{\sum_{r=1}^{4}\exp(s_{ir}/T_c)}.
\label{eq:calibration}
\end{equation}
The fitted temperature remains fixed during evaluation. The baseline uses $z_i$ as the expected score of the most probable answer under the answer-only prompt. Expected calibration error, measured in 10 equal-width bins on the 2,291 main questions, ranges from 0.018 to 0.059. We analyze the association between confidence and help-seeking separately for each model, using logistic regression of $Y_{ik}$ on the log-odds of $z_i$ with 1 intercept for each of the 54 conditions. Therefore, the coefficient gives the change in the log-odds of asking per unit change in the log-odds of confidence. We calculate 95\% percentile confidence intervals using 2,000 bootstrap resamples of questions\cite{efron1979bootstrap}. Each resampled question retains its decisions under all conditions, so contrasts between conditions are paired. Observed rates, the expected-score baseline and regression coefficients are evaluated on the same resamples.

For the experiment in Fig.~\ref{fig:coverage}, we evaluate 46 models from 9 developers on MMLU-Redux at 6 waiting costs, with 80\% stated expert accuracy and no fixed fee. The set includes open-weight models and models accessed through an application programming interface. Each model is evaluated on 100 or 1,000 questions at each waiting cost. We measure help-seeking from digit probabilities where available, otherwise using constrained choices among digits 1--5 or a prompt requesting a response in JSON format. We report model-specific help-seeking rates and confidence intervals under each model's response format and decoding settings. GPT-6 Sol and GPT-6.1 Sol are also evaluated at 95\% stated expert accuracy.

\subsection*{Consultation task and behavioral experiments}

For each multiple-choice question, models choose between answering independently and consulting an expert. Each prompt lists the scoring rules, question and answer options, consultation option, expert accuracy and consultation costs. Correct answers earn 1 point and incorrect answers 0, regardless of source, while consultation incurs the stated costs even when the expert answers incorrectly. Following the reject-option and learning-to-defer frameworks\cite{chow1970optimum,mozannar2020consistent}, we use confidence $z_i$ to estimate the expected score of independent answering. Given expert accuracy $a_H$, fixed fee $c_0$ and waiting cost $w$, the expected consultation score is $a_H-c_0-w$. Therefore, the expected-score baseline favors consultation when
\begin{equation}
z_i<a_H-c_0-w.
\label{eq:rule}
\end{equation}
Each question's confidence estimate remains fixed as the stated consultation conditions vary. Applying equation~\eqref{eq:rule} to each question gives its baseline decision, which is averaged in the same way as observed decisions.

The main experiment comprises 54 combinations of $a_H\in\{0.60,0.80,0.95\}$, $c_0\in\{0,0.1,0.2\}$ and $w\in\{0,0.1,0.2,0.45,0.8,1.5\}$. Each model receives the same questions under all conditions, with identical prompt wording apart from the stated numerical values. Let $n$ denote the number of questions and $\mathcal K(w)$ the 9 expert-accuracy and fixed-fee combinations at waiting cost $w$. The mean help-seeking rate is defined as
\begin{samepage}
\begin{equation}
h(w)=\frac{1}{9n}\sum_{i=1}^{n}\sum_{k\in\mathcal K(w)}Y_{ik},
\label{eq:help_rate}
\end{equation}
and the 2 cost responses as
\begin{equation}
\Delta_0=h(0.1)-h(0),
\label{eq:cost_onset}
\end{equation}
and
\begin{equation}
\Delta_+=h(1.5)-h(0.1).
\label{eq:cost_increase}
\end{equation}
\end{samepage}
The 2 contrasts test whether stating a cost lowers asking and whether asking falls as consultation approaches a guaranteed loss. At $w=1.5$, a correct expert answer scores at most $-0.5$ points, so the baseline never favors consultation. We classify each response by its 95\% interval $[L,U]$ and assign the first applicable category in the following order. It is a reversal when $L>0$, a decrease when $U<-0.05$ and little change when $-0.05<L$ and $U<0.05$, using an equivalence margin of 5 percentage points\cite{lakens2017equivalence}. Other responses are undecided. Persistent asking at $w=1.5$ requires the lower bound of the pooled rate to exceed 5\%.

We evaluate the 7 control variants in Supplementary Fig.~\ref{supp:controls} with stated expert accuracy fixed at 80\%. These variants assign deductions to consultation, independent answering or incorrect independent answers, vary the description of waiting costs, mention waiting without deductions, provide the expected consultation score or explicitly instruct models to maximize expected score. The 7 variants are evaluated across 6 cost levels, together with 1 additional control, yielding 43 conditions. At cost level $x$, the baseline favors consultation when $z_i<0.8-x$ for the variants involving consultation deductions, waiting wording, expected consultation scores or a goal reminder. The thresholds are $0.8+x$ for deductions on independent answering, $(0.8+x)/(1+x)$ for deductions on incorrect independent answers and 0.8 without deductions. We compare mentioning a wait without point deductions with stating only that no points are deducted. To assess the effect of extending the wait, we compare delays of 10 and 0 seconds.

We use numerical tasks without knowledge questions to measure how accurately models identify stated values, compare numbers and select the action with the higher expected score. Action selection is evaluated both from the stated accuracy and cost values and with the 2 expected scores explicitly provided. Each combination states the model's own accuracy $p$, the expert's accuracy $a_H$ and a consultation cost $w$ deducted whether or not the expert is correct. All 7 models receive the same 400 numerical combinations, sampled after removing near-ties in which the 2 expected scores differ by less than 0.05. The sample comprises 200 combinations in which consultation scores higher, 100 in which answering scores higher with $w\le1$ and 100 with $w>1$, where consultation necessarily loses points. We combine the 3 stratum-specific accuracies with weights of 1/2, 1/4 and 1/4. These weights give always-answer and always-consult strategies 50\% accuracy each on the action-selection task. Models reply with one of 2 labels, and in action selection each label denotes answering on 50\% of the trials. Accuracy thresholds are 0.95 for identifying values and comparing numbers and 0.90 for selecting actions. A threshold is met when the lower bound of the 95\% interval is at least that threshold.

Experiments on 512 held-out ARC-Challenge questions use 2 settings with 36 conditions per question. In each setting, 24 conditions combine 2 expert accuracies, 3 waiting costs including 0, neutral and waiting wording variants and 2 option-label orders. The 5 action labels are cyclically rotated by a question-specific offset to balance label preferences\cite{zheng2024selectors}, with the second mapping shifted by 2 additional positions. Another 4 conditions raise accuracy and cost together so that the expected consultation score matches that of a condition with lower expert accuracy. The remaining 8 conditions fall into 3 groups. In 2 conditions, a fixed fee replaces the waiting cost. Another 2 conditions specify a point subsidy as a negative cost. The final 4 conditions state a 0-second or longer waiting time without deducting points. The original setting uses expert accuracies of 0.80 and 0.95, waiting costs of 0.10 and 1.50 and a 10-second delay. The new setting changes the wording and uses expert accuracies of 0.70 and 0.92, waiting costs of 0.20 and 1.20 and a 12-second delay.

We add prompts that supply calibrated confidence and the expected consultation score, the expected consultation score alone or length-matched unrelated text. Comparisons include unmodified prompts, external application of the expected-score rule and independent answering on every trial. Outcomes are help-seeking decisions, independent-answer accuracy and expected scores. We calculate independent-answer accuracy from the most probable of the 4 answer options on all trials, including those ending in consultation. Behavioral contrasts use pairs of valid decisions. Comparisons of independent-answer accuracy exclude pairs with unavailable answer probabilities. Scores use the selected answer's correctness for independent answering and stated expert accuracy minus costs for consultation. If a decision fails the measurement criterion, its score is bounded by $\min(0,a_H-c_0-w)$ and $\max(1,a_H-c_0-w)$. Bounds on score changes combine the lower and upper scores of the paired prompts. We apply Holm correction separately to 2 families of 18 tests. One family tests 3 behavioral contrasts in 3 models under 2 settings; the other tests 3 prompt additions in the same models and settings.

\subsection*{Activation patching}

We conduct residual-stream activation patching in Qwen3-4B, Gemma-4-E4B-it and SmolLM3-3B using 512 held-out ARC-Challenge questions. We conduct activation-patching experiments under the original and new ARC-Challenge settings described above. Intervention sites are selected on 128 separate MMLU-Redux calibration questions under the original setting. The development scans in Supplementary Fig.~\ref{supp:swaps} cover 8 layers at the cost-number, end-of-terms and response positions on these questions. Before testing, we fix 1 layer at each of the cost-number and response positions.

For each question, we replace the residual-stream state after layer $\ell$ at token position $t$ under condition $k$ with the corresponding state under condition $k'$. Interventions are applied in both directions at the cost-number and response positions. The cost and end-of-terms positions use the last token overlapping the corresponding text span. The response position is the final token of the complete input prompt. Controls use equal-norm random perturbations and replacement of an activation with itself. For the decision margin $m$ and consultation indicator $Y$, the control-adjusted effect is defined as
\begin{equation}
\begin{aligned}
\Delta^f_{k\leftarrow k'}(\ell,t)
&=\frac{1}{2n}\sum_{i=1}^{n}\sum_{r=1}^{2}
\left[(f^P_{ikr}-f^{0,P}_{ikr})
-(f^R_{ikr}-f^{0,R}_{ikr})\right],
\qquad f\in\{m,Y\}.
\end{aligned}
\label{eq:swap_effects}
\end{equation}
Here $P$ and $R$ denote activation patching and random perturbation, respectively. Superscript $0$ denotes the unmodified output under the target condition, measured in the same batch as the corresponding intervention. The index $r$ denotes option-label order. $T$ is half the difference between the forward and reverse margin effects. Positive values indicate a greater shift toward consultation in the forward direction than in the reverse direction. Development sites are ranked by the mean absolute $T$ over the 3 behavioral contrasts, with the response-position site restricted to a relative depth of at most 0.90. Relative depth is $\ell/(N_{\ell}-1)$, where layers are indexed from 0 and $N_{\ell}$ is the number of layers. In Gemma-4-E4B-it, the last 18 of 42 layers reuse keys and values from earlier layers, so states at non-final positions from layer 23 onward are not read by later computation.

Bidirectional validation requires margin effects in both directions to agree with the signs observed in the questions used to select intervention sites and have 95\% confidence intervals excluding 0. Both directions must also have Holm-adjusted $P\leq0.05$ in two-sided, question-level tests using 2,000 sign flips across 72 directional tests. These tests cover 36 comparisons from 3 models, 2 sites, 3 contrasts and 2 settings, each in both directions. Analyses preserve the pairing of option-label orders.

\subsection*{Training objectives and evaluation}

We fine-tune Qwen3-4B, Gemma-4-E4B-it and SmolLM3-3B on MMLU-Redux using low-rank adaptation (LoRA)\cite{hu2022lora}. The 1,000-question calibration set supplies 800 training and 200 validation questions, with 2,000 separate evaluation questions. None overlap with the 2,291 main-experiment questions. Training conditions cover expert accuracies from 0.55 to 0.99, excluding the interval from 0.75 to 0.85. Waiting costs range from $-0.2$ to 0.4, and fixed fees are 0, 0.1 or 0.2. The 0.1 fee is combined only with waiting costs up to 0.2. Training prompts use neutral wording with deductions on consultation. Consultation labels rotate during training, and $m_{ik}$ is computed using the current consultation label and its 4 corresponding answer labels. Let $z_{0i}$ denote frozen base-model confidence and $A_k=a_{H,k}-c_{0,k}-w_k$ the expected consultation score. CAPT uses $q_{ik}=\sigma(m_{ik}/\tau)$ as a differentiable surrogate for the consultation decision and maximizes expected score while constraining changes to answer distributions:
\begin{equation}
\mathcal L_C=-\overline{q_{ik}A_k+(1-q_{ik})z_{0i}}+\beta(\mathrm{KL}_1+\mathrm{KL}_2)+\lambda_M(\mathcal M_1+\mathcal M_2).
\label{eq:capt}
\end{equation}
Here $\sigma$ is the sigmoid function, the overbar denotes an average over training prompts, and $\tau$ is fixed as the larger of 1 and the median absolute base-model decision margin. The 2 Kullback--Leibler (KL) divergence terms constrain answer distributions under decision and answer-only prompts, respectively. The corresponding penalties $\mathcal M_1$ and $\mathcal M_2$ constrain changes in valid-response probability mass relative to the frozen model. The weights $\beta$ and $\lambda_M$ scale the KL and probability-mass penalties, respectively. The answer KL terms measure divergence from the trained distribution to the frozen distribution, after normalization over the 4 answers. Each mass penalty is a Bernoulli KL divergence from the frozen to the trained model, using the total probability of the 5 decision digits or 4 answer digits.

We compare CAPT with SFT and soft-target training. SFT uses consultation as the target action when $A_k>z_{0i}$. Otherwise, it uses the base model's most probable answer under the same decision prompt. It retains $\mathrm{KL}_2$ and both probability-mass penalties but omits $\mathrm{KL}_1$. Soft-target training replaces the expected-score term with binary cross-entropy between $q_{ik}$ and $q^*_{ik}=\sigma[(A_k-z_{0i})/T_s]$, retaining all CAPT penalties. With the target temperature $T_s$ fixed before training at the median absolute difference between $A_k$ and $z_{0i}$, equal expected-score differences produce equal supervision targets.

All methods use rank-16 adapters on attention and multilayer perceptron projections. We select the learning rate and KL weight using validation regret from 1 seed. For each main objective, we then retrain the models with the selected settings and 3 seeds. Among checkpoints with valid decision measurements and an answer-only accuracy drop of at most 1 percentage point from the base model, we select the checkpoint with the lowest validation regret. For SmolLM3-3B SFT, no checkpoint meets the accuracy constraint, so we use the one with the lowest validation regret. Its validation answer-only accuracy is 4.0 percentage points below that of the base model. Qwen3-4B's controls with a fixed waiting cost or cyclically shifted conditions use 1 seed and CAPT-selected hyperparameters and steps. The control with a fixed waiting cost holds the waiting cost at 0.20 while varying expert accuracy and fixed fee. The control with shifted conditions cyclically shifts displayed conditions within batches while retaining the original targets.

Let $V_{ik}^{(0)}$ and $V_{ik}^{(1)}$ denote the expected scores of independent answering and consultation under the relevant cost allocation, and $Y_{iks}$ the consultation decision for seed $s$. $V_{ik}^{(0)}$ uses frozen base-model confidence $z_{0i}$, and $V_{ik}^{(1)}$ uses stated expert accuracy. Both incorporate the deductions assigned to each action and remain fixed across checkpoints. Decision regret is
\begin{equation}
r_{iks}=\max_{a\in\{0,1\}}V_{ik}^{(a)}-V_{ik}^{(Y_{iks})}.
\label{eq:regret}
\end{equation}
Regret within the training range is measured on the 2,000 evaluation questions, each under 4 conditions drawn from the training distribution. Percentage reductions compare question-averaged regret with that of the base model. Consistency is evaluated on 2 sets of conditions for the same 2,000 questions. The 8-point scan is used for $E_1$ and $E_2$. A separate set of 6 conditions forms 2 groups of 3 with matched expected consultation scores and is used for $E_3$. Let $\tilde z_{0i}$ denote $z_{0i}$ rounded down to 2 decimals. The 8 levels set the expected consultation score to $\tilde z_{0i}$ plus $-0.15$, $-0.10$, $-0.05$, 0, 0.01, 0.06, 0.11 or 0.16. The first matched group contains 3 combinations of terms with $A_k=\tilde z_{0i}-0.10$; the second contains 3 with $A_k=\tilde z_{0i}+0.11$. For an interior fitted threshold, the error is its absolute difference from $z_{0i}$. Fits at either endpoint receive a penalty of 0.20. This includes cases in which decisions never switch. Penalized threshold error $E_1$ averages these values over questions. Disagreement with the fitted threshold rule $E_2$ is averaged over questions, excluding conditions where the 2 actions have equal expected scores. Decision inconsistency $E_3$ is the average fraction of matched-score groups whose 3 decisions are not identical. We compute confidence intervals by resampling question-level values after averaging across the available training seeds. Calibration errors are evaluated separately for each seed. Candidate thresholds lie midway between adjacent scan values and 0.025 beyond either end. Fitting minimizes disagreement with observed decisions, breaking ties by proximity to $z_{0i}$ and then by the smaller threshold.

We evaluate transfer to combinations of expert accuracy and costs that are within the training range but not used in training. We also move deductions to independent answering or incorrect independent answers. Unseen combinations pair the 0.1 fee with waiting costs above 0.2 or use expert accuracies from the excluded interval. We also evaluate transfer on waiting costs from 0.4 to 0.7, with expert accuracy from 0.85 to 0.99 and no fixed fee, and on waiting-cost wording. For a deduction of $x$ points, the expected score of independent answering becomes $z_{0i}-x$ when all independent answers incur the cost and $z_{0i}-x(1-z_{0i})$ when only incorrect answers do. We compare answer-only accuracy and expected calibration error on the same questions before and after training. Calibration error uses 10 equal-width bins. The evaluation tolerances are 0.10 for $E_3$ and 0.05 for $E_1$.

\subsection*{Expert-model consultation and external routing}

We evaluate expert-model consultation on TriviaQA using 8 assistant models: Qwen3-1.7B, Qwen3-4B, Qwen3-14B, Qwen2.5-7B-Instruct, Gemma-4-E4B-it, OLMo-3-7B-Instruct, Phi-4-mini-instruct and SmolLM3-3B. Each model can answer independently or consult Qwen3-32B through the tool shown in Supplementary Fig.~\ref{supp:experttask}. After deduplication, the TriviaQA development and held-out sets contain 1,000 and 2,000 questions, respectively. Qwen3-32B answers each question once with greedy decoding. Consultation returns the stored answer, and the prompts state the expert's development-set accuracy, rounded to 64\%. We examine explicit waiting costs, server-state descriptions and battery costs charged to independent answering. Waiting times $D\in\{0,10,20,45,80,150\}$ seconds correspond to costs $w=0.01D$ points, with matching values for battery costs $x$. Crossing the 3 presentations with the 6 levels gives 18 conditions. Server-state descriptions state that each second of waiting costs 0.01 points and report the expected delay as a server status, so the model converts seconds into points. We normalize the first response line and each reference-answer alias. The response is correct if that line contains any nonempty alias. Replies that are neither an answer nor a valid tool call are scored as incorrect.

We measure verbalized confidence\cite{xiong2024uncertainty,tian2023calibration} separately by asking models to answer without tools and rate their confidence from 0 to 9. The probability-weighted mean rating is divided by 9 and calibrated using a monotonic logistic function fitted on development questions. For question $i$, external routing compares the frozen calibrated verbalized confidence $z_i$ with the expected consultation score. With stated expert accuracy $a_H=0.64$, consultation is selected when $z_i<a_H-w$ under waiting costs or $z_i<a_H+x$ under battery costs. Routing policies are compared using the same assistant answers obtained without tools and the same expert answers. Scores use stored-answer correctness minus the cost assigned to the selected action. Routing gain is the score of the external rule minus that of the model's own routing. We first average this difference over 18 conditions within each question and then calculate confidence intervals from 2,000 paired bootstrap resamples of questions. Confidence coefficients come from logistic regression of consultation on the log-odds of calibrated confidence, with 1 intercept for each condition. Sensitivity analyses use exact alias matching, rejection of responses containing negation after removal of the longest matched alias, and word-boundary matching for aliases of at most 2 characters. Each rule uses its own development-set confidence calibration. A further routing control replaces question-level confidence with development-set answer accuracy, selecting 1 fixed action per condition.

\section*{Data availability}

Model outputs and question identifiers are available in Zenodo\cite{you2026code}. The original questions and reference answers are provided by MMLU-Redux 2.0\cite{gema2025mmlu}, ARC-Challenge\cite{clark2018arc} and TriviaQA\cite{joshi2017triviaqa}.

\section*{Code availability}

The experimental and analysis code is available in the same Zenodo archive\cite{you2026code} and at \url{https://github.com/YYYJH1/llm-help-seeking}.

\section*{Funding}

This study received no funding.

\section*{Author contributions}

J.Y. designed the study, developed the code, conducted the experiments, analyzed the data and drafted the manuscript. Z.J. and C.D. contributed to study design and interpretation of the results. F.Z. and Q.W. supervised the study and revised the manuscript.

\section*{Competing interests}

No competing interests were declared.

\bibliographystyle{references}
\bibliography{refs}

\clearpage
\section*{Supplementary Information}
\setcounter{figure}{0}
\renewcommand{\thefigure}{\arabic{figure}}
\renewcommand{\theHfigure}{S\arabic{figure}}
\renewcommand{\figurename}{Supplementary Fig.}

\begin{figure}[!htbp]
\centering
\makebox[\textwidth][c]{\includegraphics[width=180mm]{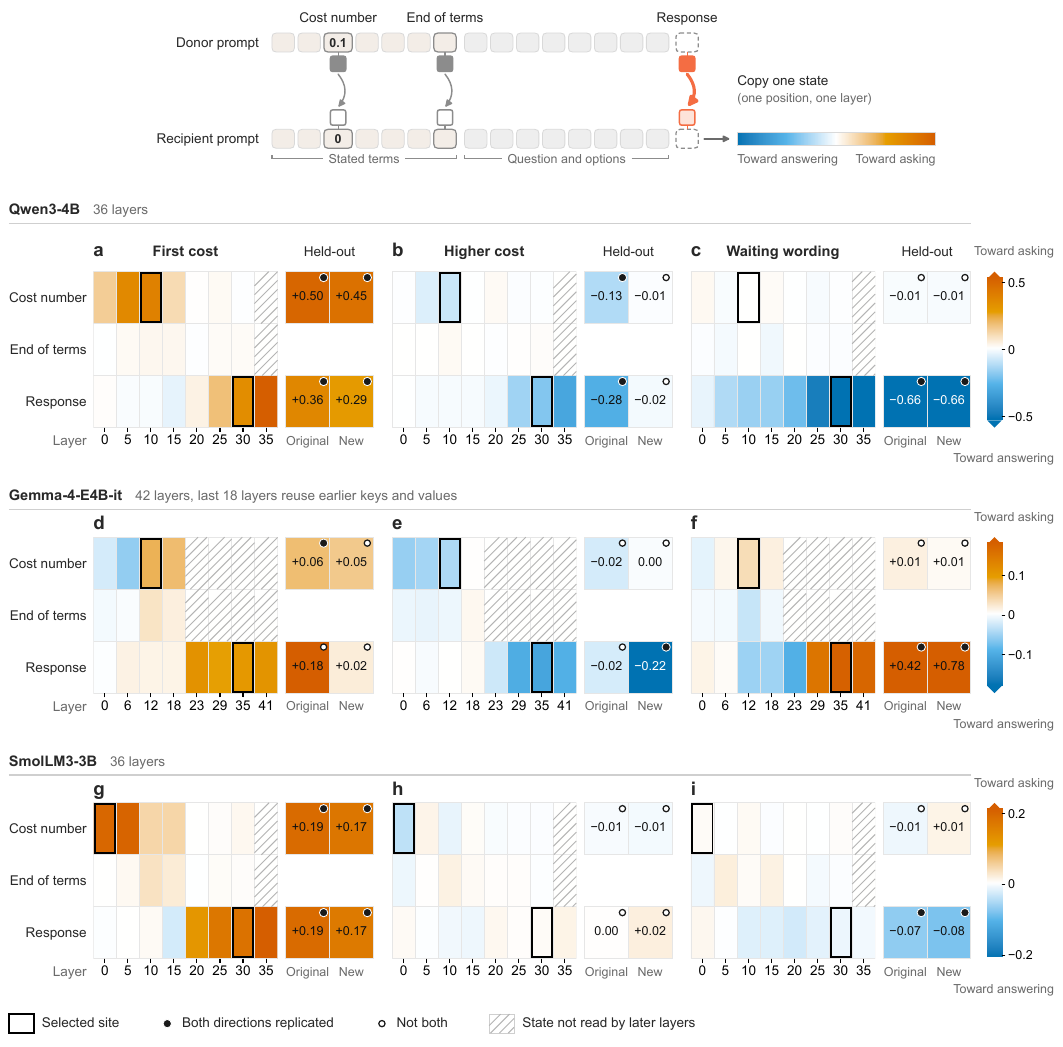}}
\caption{\textbf{Effects of activation patching across positions and layers.} States are copied between conditions for the same question at 1 layer and position. Heatmaps show scans on 128 MMLU-Redux development questions after subtracting the effects of equal-norm random perturbations. The displayed effect $T$ is half the forward-minus-reverse margin difference. Layer indices start at 0. Orange and blue indicate shifts toward asking and answering, with model-specific scales. Outlines mark 2 selected sites. Held-out columns show $T$ on 512 held-out ARC-Challenge questions under original and new settings. Filled dots mark effects in both predicted directions, using two-sided sign-flip tests with Holm-adjusted $P\leq0.05$ across 72 tests. Hatching marks states not read by later layers.}
\label{supp:swaps}
\end{figure}

\begin{figure}[!tbp]
\centering
\makebox[\textwidth][c]{\includegraphics[width=180mm]{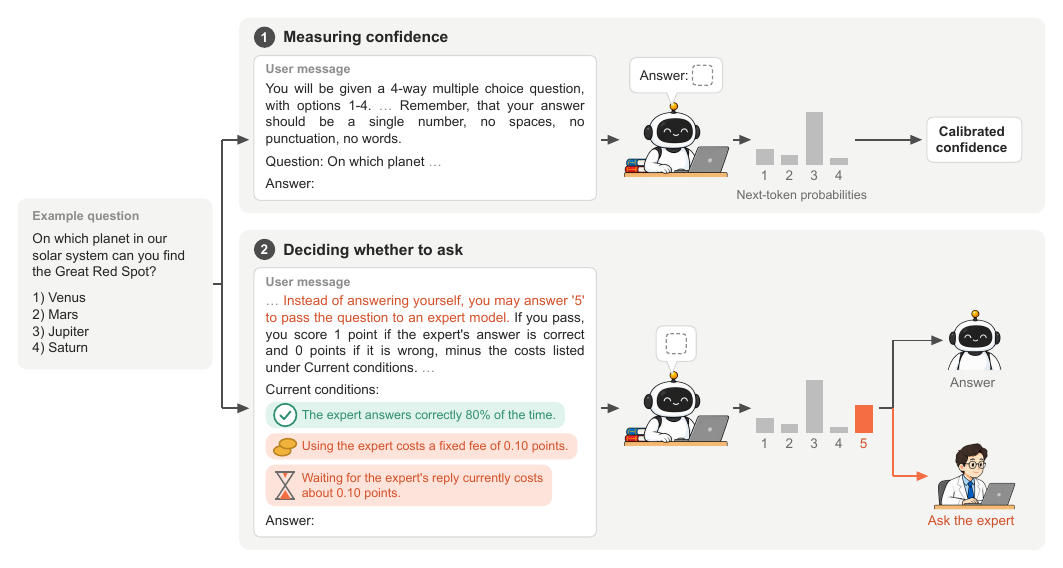}}
\caption{\textbf{Prompts for confidence measurement and help-seeking.} Both stages use the same MMLU-Redux calibration question. Stage 1 measures confidence with an answer-only prompt and the fixed prefix ``Answer: ''. Stage 2 adds expert accuracy, fixed fee and waiting cost for one of 54 conditions. Bars illustrate next-token probabilities. Digits 1--4 select answers, and digit 5 requests expert help.}
\label{supp:prompts}
\end{figure}

\begin{figure}[!tbp]
\centering
\makebox[\textwidth][c]{\includegraphics[width=180mm]{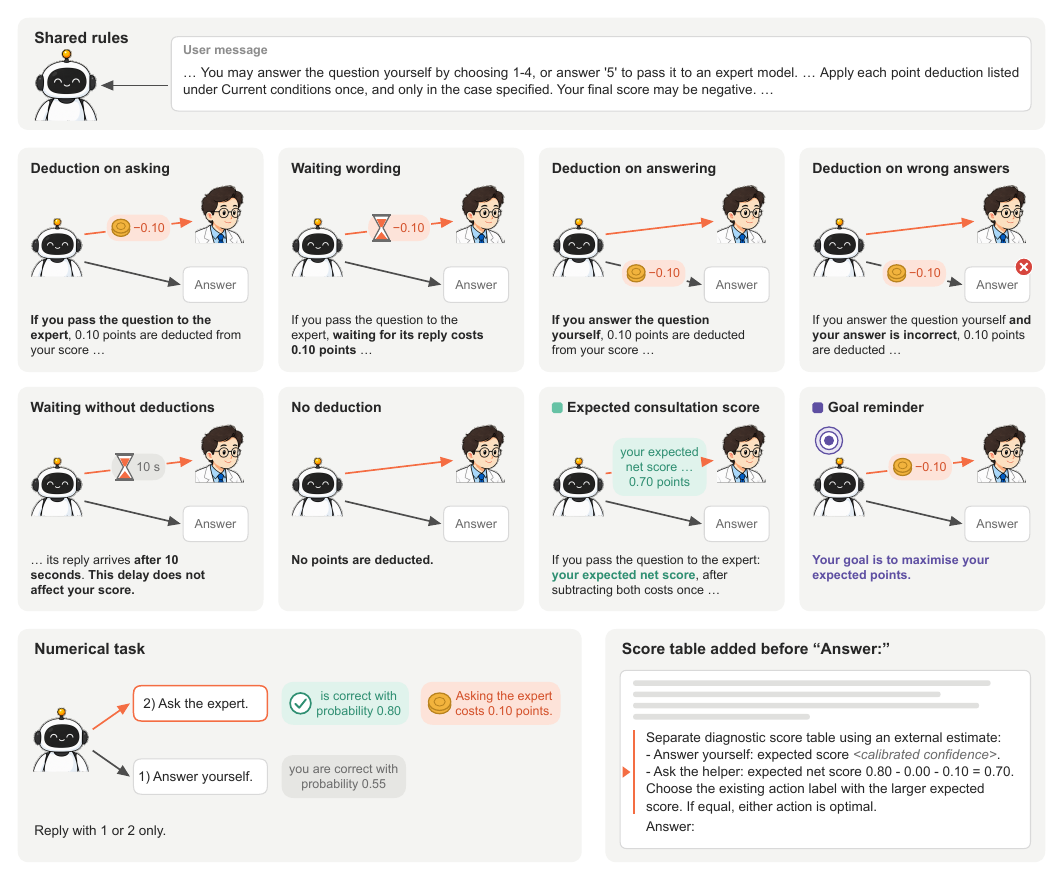}}
\caption{\textbf{Prompts in the control experiments.} Of the 8 control prompts, 6 share the scoring rules shown at the top. The expected-score prompt and goal reminder use the main task rules. Tiles show cost allocation and wording. Coins indicate 0.10-point deductions, hourglasses stated waits, and crosses deductions for incorrect independent answers. The lower panels show the numerical action-selection prompt and the score table inserted before the final ``Answer:''.}
\label{supp:controls}
\end{figure}

\begin{figure}[!tbp]
\centering
\makebox[\textwidth][c]{\includegraphics[width=180mm]{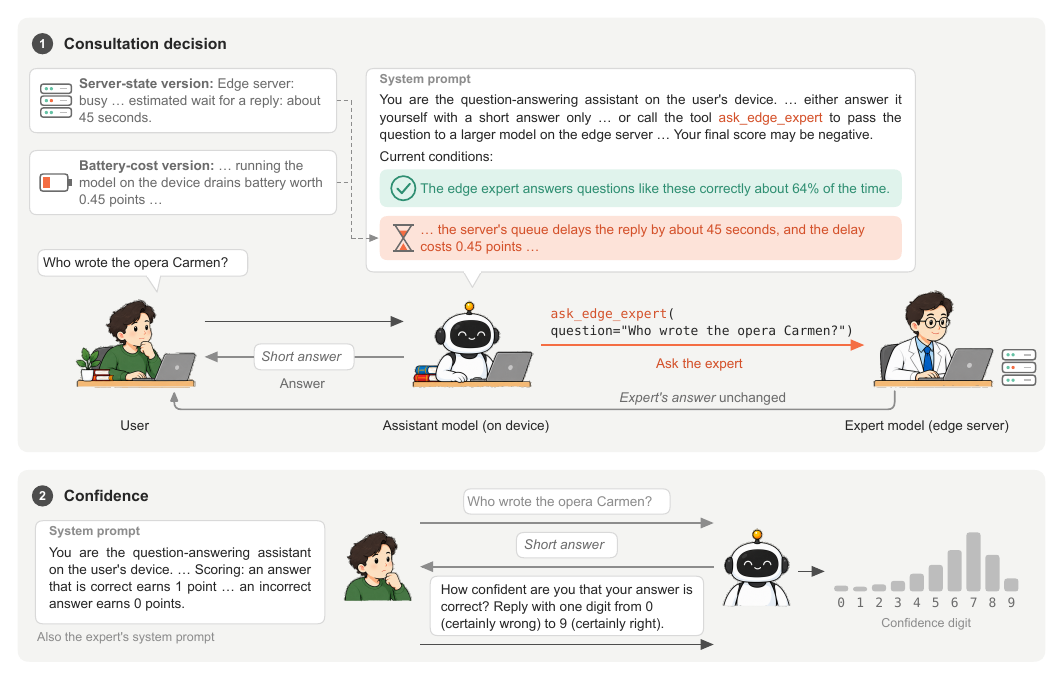}}
\caption{\textbf{Prompts in the expert-model task.} Stage 1 states expert accuracy and cost in the system prompt. The assistant answers independently or calls \texttt{ask\_edge\_expert}. The tool returns the expert's answer unchanged. Upper-left cards show server-state and battery-cost variants. Server-state rules charge 0.01 points per second. Stage 2 measures confidence separately using an answer-only system prompt and a follow-up 0--9 rating. The expert uses the same answer-only system prompt.}
\label{supp:experttask}
\end{figure}

\end{document}